\documentclass{elsart}

\usepackage[T1]{fontenc}    
\usepackage{hyperref}       
\usepackage{url}            
\usepackage{booktabs}       
\usepackage{amsfonts}       
\usepackage{nicefrac}       
\usepackage{microtype}      
\usepackage{xcolor}
\usepackage{graphicx}
\usepackage{amsmath}
\usepackage[utf8]{inputenc} 

\begin{document}
\begin{frontmatter}
	\title{PRIS: Practical robust invertible network for image steganography}
	\author{Hang Yang$^1$},
	\author{Yitian Xu$^1$*}
	\ead{xytshuxue@126.com, Tel.: +8610 62737077.}
	\author{Xuhua Liu$^1$*}
	\author{Xiaodong Ma$^1$*}
	\address{$^1$College of Science, China Agricultural University, Beijing 100083, China}

	\begin{abstract}
		Image steganography is a technique of hiding secret information inside another image, so that the secret is not visible to human eyes and can be recovered when needed. Most of the existing image steganography methods have low hiding robustness when the container images affected by distortion. Such as Gaussian noise and lossy compression. This paper proposed PRIS to improve the robustness of image steganography, it based on invertible neural networks, and put two enhance modules before and after the extraction process with a 3-step training strategy. Moreover, rounding error is considered which is always ignored by existing methods, but actually it is unavoidable in practical. A gradient approximation function (GAF) is also proposed to overcome the undifferentiable issue of rounding distortion. Experimental results show that our PRIS outperforms the state-of-the-art robust image steganography method in both robustness and practicability. Codes are available at \url{https://github.com/yanghangAI/PRIS}, demonstration of our model in practical at \url{http://yanghang.site/hide/}.
	\end{abstract}
	
\end{frontmatter}
	
\section{Introduction}
\label{sec:intro}
Steganography is the art of concealing information. The goal of steganography is to hide a secret message within a host medium \cite{Kessler2011Overview,Johnson1998Exploring}. The host medium containing the hidden secret message is called the container \cite{lu2021large}. The container is typically publicly visible, but the difference between the host and container should be invisible to third parties. There are some good introductions to steganography in \cite{Kessler2011Overview,kessler2004overview,fridrich2011breaking}. Image steganography aims to hide message such as image, audio, and text within a host image in an undetectable way \cite{RIIS}, meaning the host medium in image steganography is an image. Nowadays, image steganography is widely used in fields such as copyright protection, digital communication, information certification, and more \cite{cheddad2010digital}.

Traditional image steganographic methods have the ability to hide only a small amount of information \cite{barni2001improved,fridrich2001detecting,hsu1999hidden,lerch2016unsupervised,luo2010edge,ruanaidh1996phase}. They conceal secret messages in the spatial or adaptive domains \cite{kadhim2019comprehensive} with capacities of 0.2$\sim$4 bits per pixel (bpp). In recent years, some researchers have attempted to hide an image containing more information than the secret message in traditional image steganographic methods within a host image \cite{Baluja2017Hiding,zhang2020udh,baluja2019hiding,duan2019reversible,duan2020high,duan2020highx}. They introduced deep learning into image steganography by using two separate networks to realize the embedding and extraction processes. Lu \cite{lu2021large} and Jing \cite{jing2021hinet} achieved state-of-the-art performance in image steganography by using an invertible network to embed a secret image into a host image. Since the extraction process is the inverse of the embedding process, the secret image can be fully recovered.

In practical applications, the container image is often subject to various forms of attacks due to factors such as lossy image compression for storage savings. These attacks can distort the container image, potentially affecting the ability to extract the secret information. Robustness refers to the ability of the secret information to remain unchanged when the container is distorted \cite{byrnes2021data}. In other words, it measures how similar the secret and extracted messages will be when the container image is attacked.

However, state-of-the-art image steganography methods \cite{lu2021large,jing2021hinet} did not take into account the effect of container images being attacked. As a result, they failed to extract secret images when container images were distorted. To address this issue, Xu \cite{RIIS} proposed robust invertible image steganography. This method improved the robustness of image steganography by taking image distortion into account, allowing for the successful extraction of secret information even when the container image is distorted.

Moreover, since the current mainstream deep learning frameworks have a numerical precision of 32 bits, while images are usually 8 bits, there is an inevitable rounding error when saving the container images. However, to the best of our knowledge, the existing deep steganography methods ignore this problem \cite{wan2022comprehensive,RIIS,jing2021hinet,lu2021large}, which is not practical in real application, we also give a method to illustrate the importance to consider rounding error in section 3.7.

In this paper, we designed a practical robust invertible network called PRIS based on HiNet \cite{jing2021hinet}. We introduce two enhancement modules and a 3-step training strategy into invertible neural network to improve its robustness. In addition, we take rounding distortion into account and propose a gradient approximation function (GAF) to deal with the undifferentiable problem of rounding operation. The main contributions are listed as follows:

1. A practical robust invertible network is proposed for image steganography under diverse attacks.

2. We introduce a 3-step training strategy into our training process to achieve a better robustness.

3. A gradient approximate function is proposed to solve the undifferentiable problem caused by rounding operation, and take rounding error into consideration.

4. Experiments results demonstrate that our proposed PRIS outperforms the existing state-of-the-art method RIIS in both robustness and practicability.

\section{Related work}
\subsection{Traditional image steganography}
Traditional image steganography can be divided into two categories based on the whether the hiding process happen in spatial or frequency domain. \textbf{Spatial domain:} The most popular spatial-based method called Least Significant Bit (LSB) \cite{chan2004hiding,tamimi2013hiding}, it changes $n$ least significant bits of host image to embed secret messages. However, the change of picked bits make it easy to be detected by some steganalysis methods \cite{fridrich2001detecting,hawi2004steganalysis,zhi2003lsb}. In addition, Pan \cite{pan2011image} utilizes pixel value differencing (PVD), Tsai \cite{tsai2009reversible} introduces histogram shifting, Nguyen \cite{nguyen2006multi} use multiplebit-planes and Imaizumi \cite{imaizumi2014multibit} propose palettes in image steganography. \textbf{Frequency domain: }Those methods hide secret messages in frequency domains, such as discrete cosine transform (DCT) \cite{hsu1999hidden}, discrete Fourier transform (DFT)  \cite{ruanaidh1996phase}, and discrete wavelet transform (DWT) \cite{barni2001improved} domains. Although they are more robust and undetectable than LSB methods, they still suffered from limited payload capacity.
\subsection{Deep learning-based image steganography}
Recently, many researchers have applied deep learning methods to image steganography and achieved better performance than traditional methods. HiDDeN \cite{zhu2018hidden} and SteganoGAN \cite{zhang2019steganogan} utilize the encoder-decoder architecture to realize the embedding and extraction of secret messages and employ a third network to resist steganalysis adversarially. Shi \cite{shi2018ssgan} proposes Ssgan for image steganography, which is based on generative adversarial networks. Baluja \cite{baluja2019hiding,Baluja2017Hiding} and Zhang \cite{zhang2020udh} hide a secret image of the same size as the host image with deep learning method, achieve a much higher payload capacity than traditional methods.
\subsection{Invertible neural network}
Dinh \cite{dinh2014nice} proposed Invertible neural network (INN) in 2014. It learns a stable bijective mapping between a data distribution $p_X$ and a latent distribution $p_Z$. Unlike CycleGAN \cite{zhu2017unpaired}, which uses two generators and a cycle loss to achieve bidirectional mapping, INN performs both forward and backward propagation within the same network, acting as both feature encoder and image generator.
INNs are also useful for estimating the posterior of an inverse problem \cite{ardizzone2018analyzing}. More flexible INNs are built with masked convolutions under some composition rules in \cite{song2019mintnet}. An unbiased flow-based generative model is proposed in \cite{chen2019residual}. Other works improve the coupling layer for density estimation, such as Glow \cite{kingma2018glow} and i-ResNet \cite{behrmann2019invertible}, resulting in better generation quality.

Lu \cite{lu2021large}, Jing \cite{jing2021hinet} and Jia \cite{jia2023afcihnet} introduced INN into image steganography. The strict invertibility of INN just meets the requirement that embedding and extraction are mutually inverse processes. Therefore they gain state-of-the-art performance in image steganography.  However, the strict invertibility also means that when the container image is attacked, the noise is also transmitted to the extracted image, which make it vulnerable to attack. Xu \cite{RIIS} proposed robust invertible image steganography (RIIS) and improve the robustness of the aforementioned invertible neural networks.

In the field of image steganography, the invertible network has shown its great potential. It achieves state-of-the-art performance by utilizing the prior knowledge that hiding and extraction are inverse processes \cite{lu2021large,jing2021hinet}, along with the strict reversibility of its own network. However, due to its strictly reversible nature, when the container image is perturbed, the inverse process of its network will also be perturbed. This implies that the extracted image will also be distorted. To address this issue, we proposed our model PRIS, by introducing two new modules: pre-enhance and post-enhance, which are added before and after the extraction process respectively, and through a 3-step training strategy, we improve its robustness to state-of-the-art level. Moreover, rounding error is considered in our PRIS since it is unavoidable in practice, and a gradient approximate function is proposed to address the undifferentiable problem of rounding operation.
\section{Method}
\subsection{Overview}
\begin{figure*}[!htbp]
	\centering
	\includegraphics[width=0.7\textwidth]{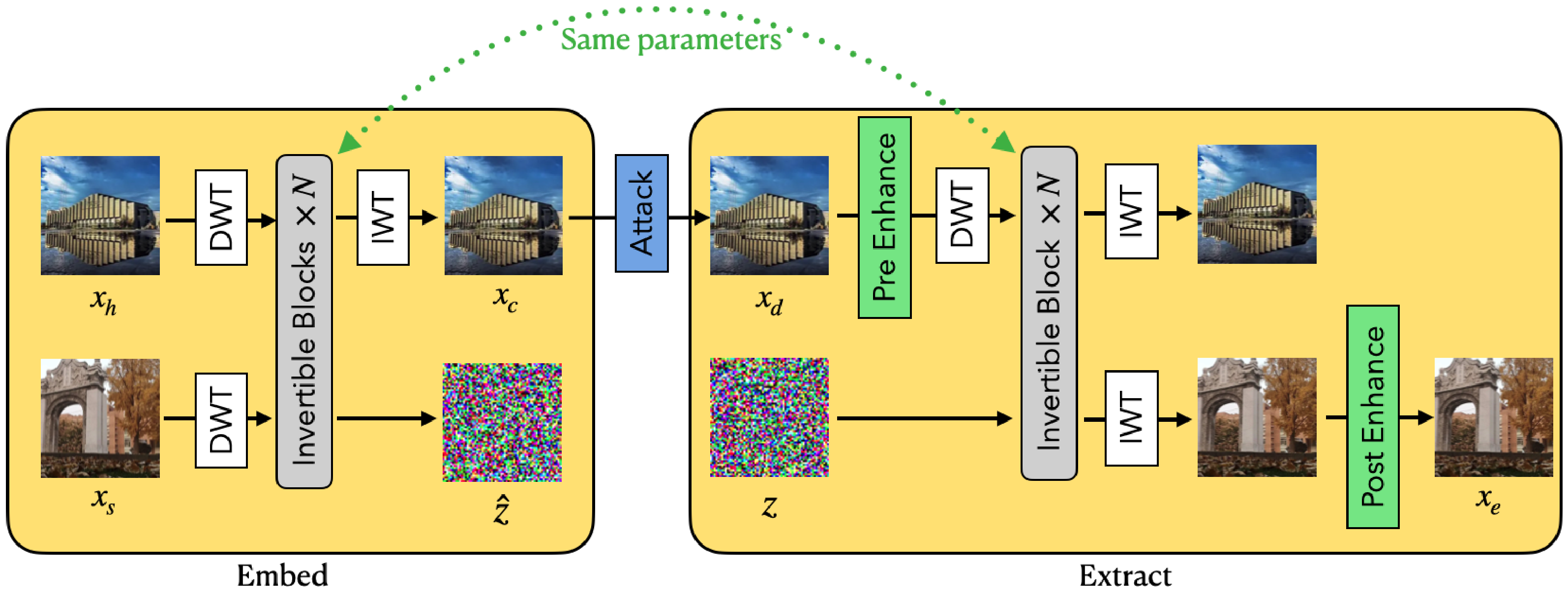}
	\caption{\textbf{The framework of PRIS}, it contains two main parts, invertible blocks and enhance modules, DWT and IWT denote discrete wavelet transform and inverse wavelet transform respectively. The left block is the embedding process and the right block is the extraction process. $z$ is a random noise that follows a normal distribution.}
	\label{overview}
\end{figure*}

Let $x_{h}$, $x_{s}$, $x_{c}$ and $x_{e}$ denote the host, secret, container and extracted image respectively.
The architecture of our model is shown as Fig. \ref{overview}. In the embedding process, the $x_h$ and $x_s$ are first transformed to the frequency domain by discrete wavelet transform (DWT), and then fed into $N$ invertible blocks, which output two images. The output corresponding to image $x_h$ is further transformed to the spatial domain by IWT, resulting in $x_c$. The output $\hat{z}$ corresponding to image $x_s$ is inaccessible in the extraction process, and we hope that it follows a Gaussian distribution, so that we can input a Gaussian distribution in the extraction process. In practical, $x_c$ will be attacked by different distortion and becomes image $x_d$.
In the extraction process, if pre-enhance module is enabled, $x_d$ is enhanced by pre-enhance first and then transformed by DWT before being sent to the inverse process of invertible blocks. Meanwhile, a Gaussian distributed image $z$ is also sent to the invertible blocks. Finally, we obtain two images, revealed host and extracted secret in the frequency domain. The latter is transformed to the $x_e$ by inverse wavelet transform (IWT). If post-enhance is enabled, this image will be further enhanced by a post-enhance module.

\subsection{Invertible block}
\begin{figure*}[!htbp]
	\centering
	\includegraphics[width=0.75\textwidth]{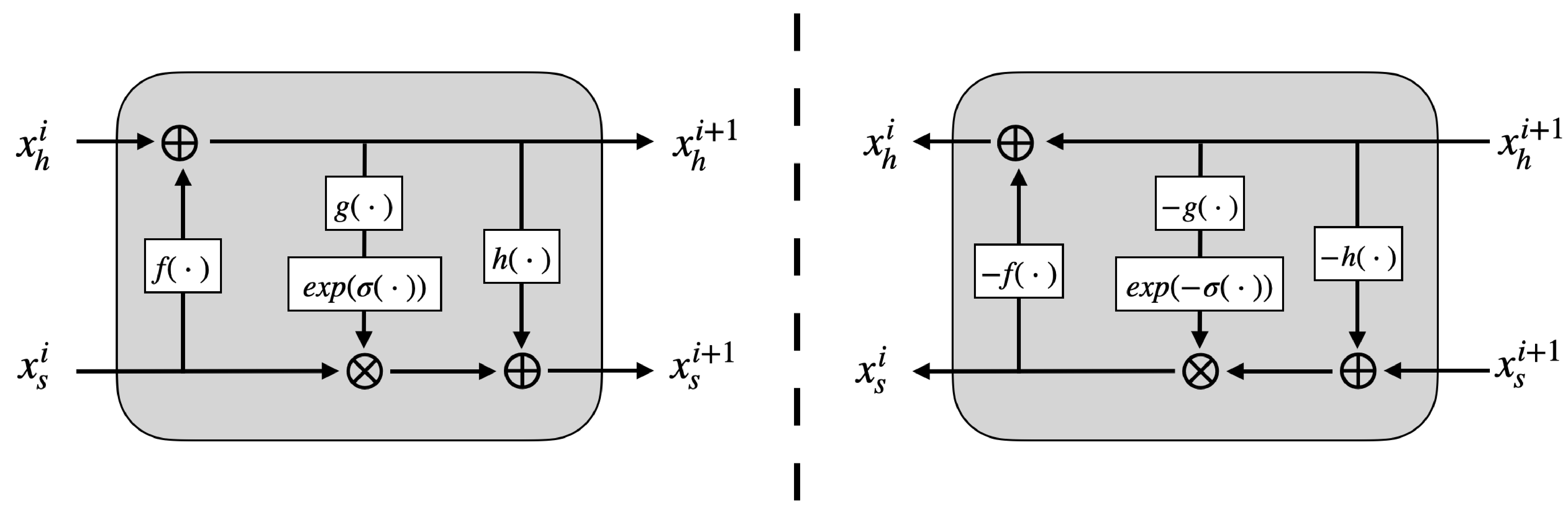}
	\caption{\textbf{Network architecture of an invertible block.} The left part shows the forward process, i.e., the embedding process. The right part shows the backward process, i.e., the extraction process. $f(\cdot)$, $g(\cdot)$ and $h(\cdot)$ represent three convolutional networks that share the same architecture, $\sigma$ denotes sigmoid activate function, $\otimes$ and $\oplus$ denote element-wise multiply and add.}
	\label{INblock}
\end{figure*}

Fig. \ref{INblock} shows the overview of invertible block. In the forward process, it takes $x_h^{i}$ and $x_s^{i}$ as input, and outputs $x_h^{i+1}$ and $x_s^{i+1}$ with the same size as the input. Eq. \eqref{ibf2} gives the formula of it.
\begin{align}
	x_h^{i+1} &= x_h^i + f(x_s^i) \nonumber \\	
	x_s^{i+1} &= x_s^i \otimes exp(\sigma(g(x_h^{i+1}))) + h(x_h^{i+1}) \label{ibf2}
\end{align}
By transforming Eq. \eqref{ibf2}, we can easily obtain their inverse processes. The result is shown as below:
\begin{align}
	x_s^i &= (x_s^{i+1}-h(x_h^{i+1})) \otimes exp(-\sigma(g(x_h^{i+1}))) \nonumber   \\
	x_h^i &= x_h^{i+1}-f(x_s^{i}) \label{ibb2}
\end{align}
The invertible block takes two images as input and outputs two images for both forward and inverse processes, as implied by Eqs. (\ref{ibf2}) and (\ref{ibb2}). However, in image steganography, only one container image is transferred in communication. To address this issue, an auxiliary variable $z \sim N(0, 1)$ is taken as the second image and fed into the inverse process with the container image $x_c$.

\subsection{Enhance module}

We introduced pre-enhance and post-enhance modules at the beginning and end of the backward process respectively. Both modules share the same architecture of densenet \cite{huang2017densely}.
The pre-enhance module takes the distorted container image as input and outputs an enhanced container image that facilitates the extraction by the invertible block. The post-enhance module takes the extracted image as input and outputs an enhanced extracted image that is more similar to the original secret image.

The pre-enhance module aims to reduce the perturbation of the container image before feeding it into the extraction network, while the post-enhance module aims to restore the quality of the extracted image after getting it from the extraction network. In this way, we can mitigate the negative effects of perturbation on both ends of the extraction process and enhance the robustness of our method.

Meanwhile, the pre-enhance and post-enhance modules can also weaken the strict reversibility of the invertible network, so that it can better resist noise. We use the analogy of a stiff stick and a soft rope to illustrate the relationship between the strict reversibility and the noise transmission. A stiff stick will transmit all the perturbations from one side to the other, while a soft rope will dampen most of the perturbations from the other side. Therefore, we introduce the enhance modules to relax the strict reversibility of INNs. By weakening the reversibility of the network, we can reduce the impact of noise on both ends of the extraction process and enhance the robustness of our method.

\subsection{Loss function}
We are concerned with two types of similarity: one is the similarity between the host image $x_h$ and the container image $x_c$, i.e., c-pair; the other is the similarity between the secret images $x_s$ and the extracted images $x_e$, i.e., s-pair. We use the PSNR (Peak Signal to Noise Ratio) \cite{setiadi2021psnr} metric to measure this similarity. Higher PSNR means higher similarity. PSNR-C and PSNR-S denote the PSNR values of c-pair and s-pair, respectively. Our goal is to maximize PSNR-C and PSNR-S. To achieve this, we introduce the following two losses, $L_c$ and $L_s$:
\begin{align}
L_c= \sum_{p}(x_c^{(p)}-x_h^{(p)})^2, L_s= \sum_{p}(x_s^{(p)} - x_e^{(p)})^2
\end{align}
where $x^{(p)}$ represents the pixel value at position $p$ for image $x$, and $L_c$ and $L_s$ measure the similarity of c-pair and s-pair respectively. The total loss is a weighted sum of the two losses:
\begin{equation}
L = \lambda_cL_c+\lambda_sL_s
\end{equation}
There is a trade-off between $L_c$ and $L_s$. By increasing $\lambda_c$, we give more influence to $L_c$ in the total loss, which improves the PSNR-C at the expense of reducing the PSNR-S. Conversely, we can enhance the PSNR-S by sacrificing some PSNR-C by increasing $\lambda_s$.
\subsection{3-step training strategy}

The strict invertibility is vulnerable to attack, but meets the demand of image steganography, therefore, we need to find a balance point between strict invertiblity and non-invertiblity. The existence of enhance modules breaks too much invertiblity,  in order to balance it, we proposed 3-step training strategy. It splits whole training process into 3 steps. 

{\bf Step 1:} Pre-train invertible blocks. In this step, only the invertible blocks are enabled, the enhance modules are not computed both in forward and backward process. This will guarantee the invertiblity of whole networks. 

{\bf Step 2:} Pre-train enhance modules. In this step, all parts of the PRIS are enabled, but the parameters of invertible block are frozen, by doing this, it will weaken the invertiblity, while make sure it will not damage the performance.

{\bf Step 3:} Finetune invertible blocks and enhance modules together. In this step, all parameters are enabled and will be updated during the back-propagation. It will help to improve the robustness furthermore.

\subsection{Why rounding error is so important?}
\begin{figure}[htbp]
	\centering
	\includegraphics[width=0.85\linewidth]{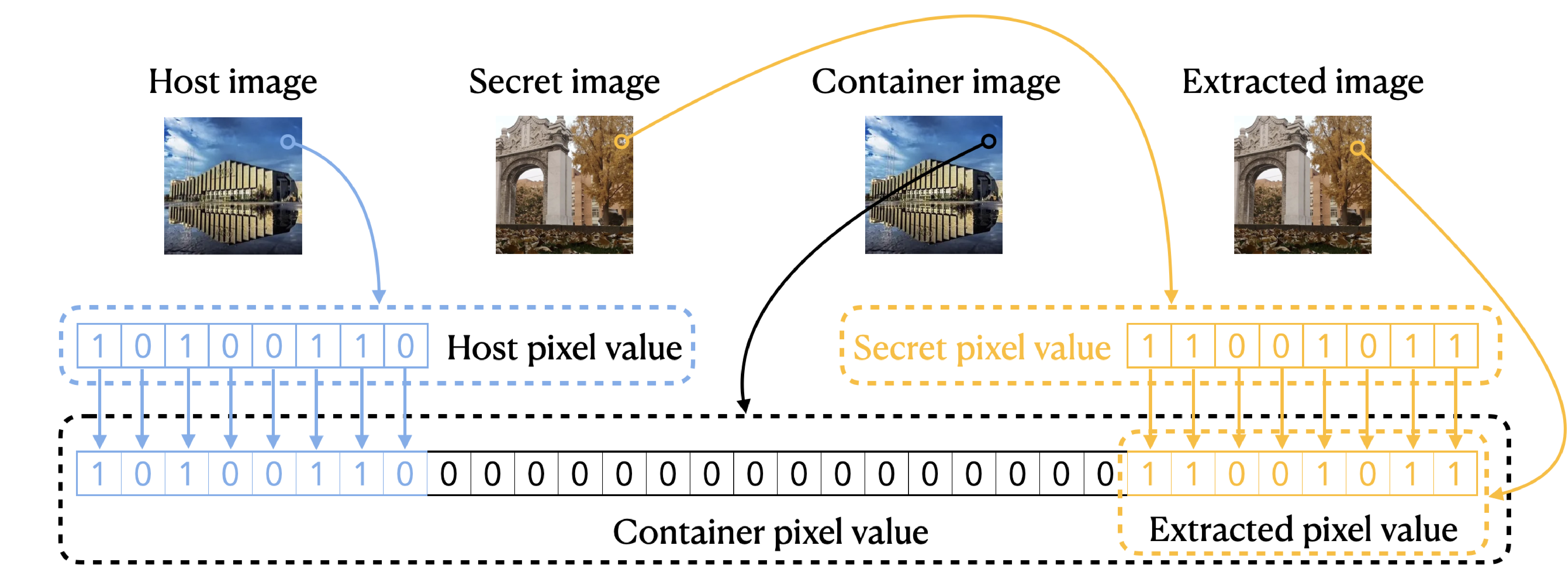}
	\caption{The method to hide and extract if the rounding error is ignored. All the pixel value is shown in binary way.}
	\label{rounderr}
\end{figure}

In this section, we present a method to demonstrate that we can embed an 8-bit secret image into an 8-bit host image and obtain a 32-bit container image with a guaranteed PSNR-C of more than 144.52dB and a lossless extraction if we neglect the rounding error. The overview of this method is shown as Fig.\ref{rounderr}.
Let $x_{hi}$, $x_{si}$, $x_{ci}$ and $x_{ei}$ denote the i-th pixel of host, secret, container and extracted image respectively.
$x_{hi}$, $x_{si}$ and $x_{ei}$ are integers ranging from 0 to $2^8-1$, and $x_{ci}$ is an integer ranging from 0 to $2^{32}-1$. The formulas to compute $x_{ci}$ and $x_{ei}$ are given by Eqs. (\ref{hideround}) and (\ref{extractround}). It is clear that $x_{ei}=x_{si}$, so the extraction is a lossless process.
\begin{align}\label{hideround}
	x_{ci} &=2^{24}  \cdot x_{hi} + x_{si}   \\
	\label{extractround}
	x_{ei} &= x_{ci} -  2^{24}\cdot \lfloor x_{ci} \cdot 2^{-24} \rfloor
\end{align}
Moreover, the PSNR-C is greater than 144.52dB according to Eqs. (\ref{mse}) and (\ref{psnr}).
\begin{align}\label{mse}
	MSE&=\frac{1}{N}\sum_{i=1}^N(2^{24}\cdot x_{hi} - x_{ci})^2=\frac{1}{N}\sum_{i=1}^N x_{si}^2\leq 255^2\\
	\label{psnr}
	PSNR &=10\cdot \log_{10}\frac{MAX^2}{MSE} \geq 10\cdot \log_{10}\frac{(2^{32}-1)^2}{255^2} > 144.52
\end{align}
Where $N$ denotes the total pixels number of the host (secret) image.
\subsection{Differentiable rounding function}
The rounding function $round$ has a zero gradient almost everywhere, which leads to gradient vanishing during the training process. To address this issue, we tried two ways to replace the gradient of $round$. First, we used 1 as the gradient of $round$, which means during the backward stage, we computed the gradient with $y=x$. Second, we proposed a modified gradient approximation function (GAF), defined by Eqs. (\ref{roundapp1}) and (\ref{roundapp2}).
\begin{align}
	\begin{split}
		sign(x)= \left \{
		\begin{array}{ll}
			1,  & if \; \lfloor x \rfloor \; is \; odd\\
			-1, & else
		\end{array}
		\right.
	\end{split}
	\label{roundapp1}
\end{align}
\begin{equation}
	GAF(x)=sign(x)\cdot 0.5 \cdot \cos(\pi \cdot x) + 0.5 + \lfloor x \rfloor
	\label{roundapp2}
\end{equation}
We designed GAF based on the following two features that it shares with the rounding function:

1. It acts as an identity mapping when the input is an integer, so the gradient is zero.

2. The gradient reaches its maximum value when the input is at the midpoint between two consecutive integers.
\begin{figure}[!htbp]
	\centering
	\includegraphics[width=0.5\linewidth]{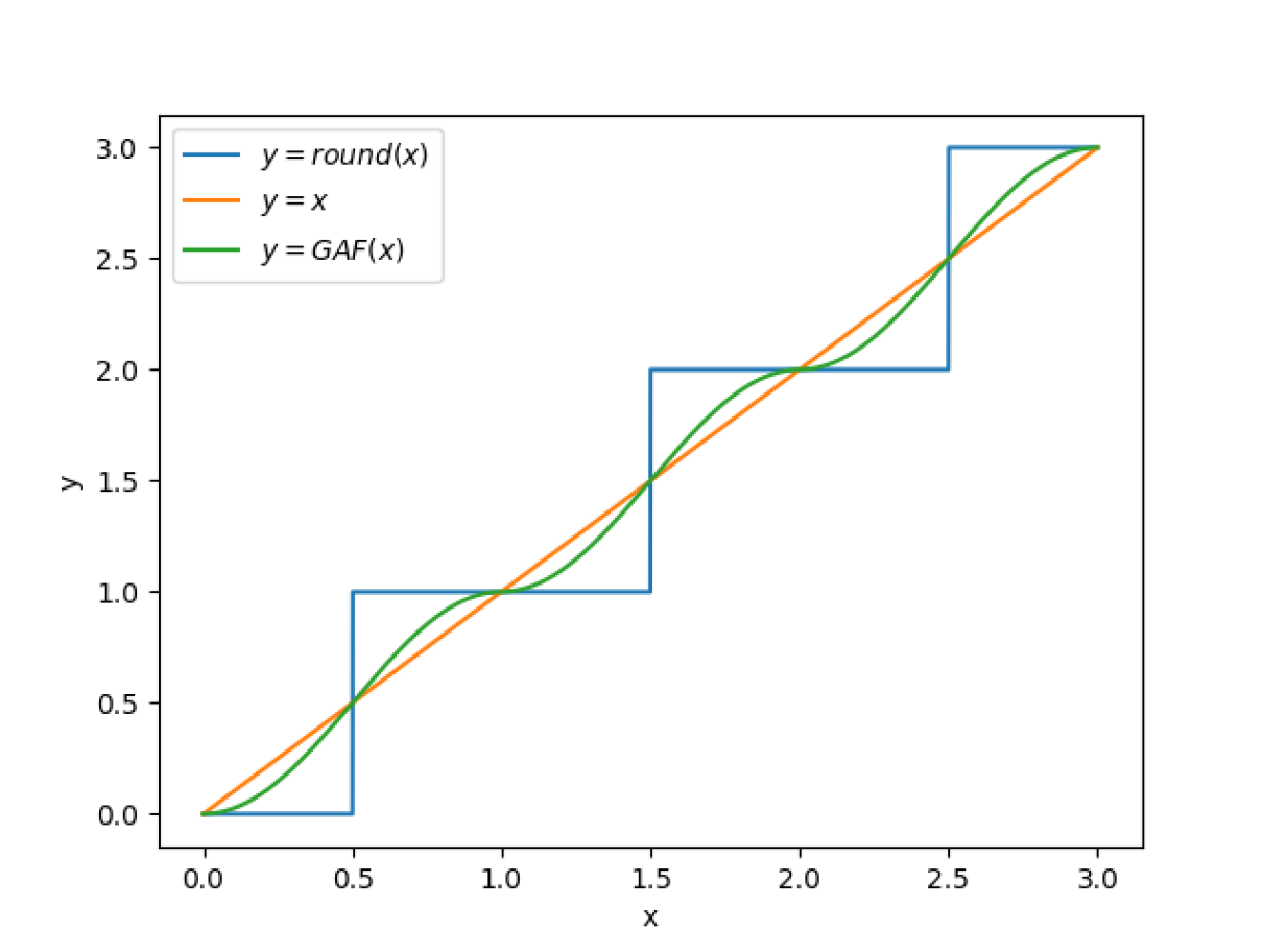}
	\caption{Round and its approximate function.}
	\label{round}
\end{figure}
\section{Experiment}
\subsection{Implementation details}
We train our model on the DIV2K training dataset \cite{agustsson2017ntire} and evaluate it on the DIV2K testing dataset on RTX3090. The input images are cropped to a resolution of 224 $\times$ 224 pixels. For the training dataset, we apply random cropping for better generalization. For the testing dataset, we use center cropping to avoid randomness in evaluation. Each stage consists of 1600 epochs and we use the Adam optimizer with $\beta_1=0.9$ and $\beta_2=0.99$. The initial learning rates are $10^{-4.5}$ and  $10^{-5.5}$ for steps 1, 2 and step 3 respectively, reducing them by half every 200 epochs for all steps. We set $\lambda_c$ and $\lambda_s$ to 1 unless otherwise specified.
\subsection{Ablation experiments}
{\bf Effectiveness of enhance module, 3-step training strategy and domain selection.} As shown in the second and third rows of Table \ref{ablation}, both pre-enhance and post-enhance modules improved PSNR-C and PSNR-S compared to the baseline in the first row. Moreover, applying both modules together achieved further improvement. This indicates that the enhance modules can effectively reduce the noise of the distorted container image before and after the inverse process of the invertible blocks. The pre-enhance module can help the invertible blocks to resist the noise from distortion, while the post-enhance module can further refine the extracted image and remove some residual noise.
\begin{table}[!htbp]
	\centering
	\caption{Ablation studies for every model design, including pre-enhance, post-enhance, 3-step strategy and domain selection in PRIS under Gaussian $\sigma=10$ distortion.}
		\begin{tabular}{ccccccc}
		\toprule
		3-step & Pre-en. & Post-en.  & Domain & PSNR-C  & PSNR-S            \\ \midrule
		-&$\times$ & $\times$ & - & 30.79 & 28.99   \\
		$\checkmark$&$\times$ & $\checkmark$  & Spatial & 31.02 & 29.24   \\
		$\checkmark$&$\checkmark$ & $\times$  & Spatial & 31.11 & 29.28    \\
		$\times$&$\checkmark$ & $\checkmark$ & Spatial & 30.97 & 27.63     \\
		$\checkmark$&$\checkmark$ & $\checkmark$ & Frequency & 31.09 & 29.03     \\
		$\checkmark$&$\checkmark$ & $\checkmark$ & Spatial & \textbf{31.13} & \textbf{29.34}     \\
		\bottomrule
		\label{ablation}
	\end{tabular}
\end{table}
{\bf Spatial domain or frequency domain?}

Junpeng Jing et al. \cite{jing2021hinet} applied discrete wavelet transform (DWT) to convert images from spatial domain to frequency domain before sending them to the invertible block. Then, they used inverse wavelet transform (IWT) to convert images back from frequency domain to spatial domain after the invertible block. This means that the invertible blocks processed images in frequency domain. Their research demonstrated that hiding images in frequency domain is more effective than in spatial domain. However, weather frequency domain still the better option for enhance modules needs to be checked, therefore we did the ablation study of different domains. 

 The comparison between the fifth and sixth rows of Table \ref{ablation} shows that the spatial domain has a better performance than the frequency domain. Therefore, the enhance modules will process images in the spatial domain in other experiments. This may be because the spatial domain is more suitable for capturing the local features and edges of the image, which are important for image restoration. The frequency domain, on the other hand, may introduce some artifacts or distortions when transforming the image back to the spatial domain.

\begin{figure}[!htbp]
	\centering
	\includegraphics[width=0.6\linewidth]{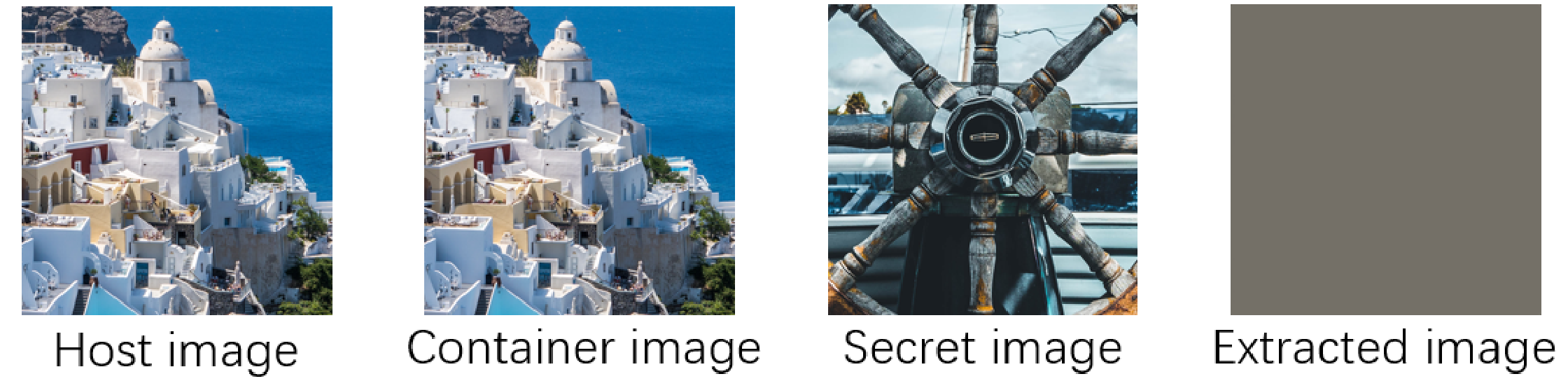}
	\caption{The visual result of setting the gradient of rounding function to 0, it achieves a high PSNR-C by ignoring extraction task, essentially, it tries to learn an identical mapping function.}
	\label{rounddiff0}
\end{figure}

{\bf Different ways to compute the gradient of rounding function.} We experimented with three different ways to compute the gradient of rounding function – setting all gradients to 0, setting all gradients to 1, and replacing $round$ with GAF during backward propagation. Table \ref{round} shows the results of each method. When we set the gradient of rounding function to 0, it means that the loss of secret-pair will not propagate to the embedding process, and thus it will not take extraction into account. The visual results in Fig. \ref{rounddiff0} demonstrate that it completely ignores the extraction task, and essentially, the goal becomes learning an identical mapping function. From Table \ref{round}, we can see that GAF is generally better than other methods. Therefore, we choose GAF to compute the gradient. The GAF can provide a smooth approximation of the rounding function and preserve more information during backpropagation, which can contribute to the learning process of the network.
\begin{table}[!htbp]
	\centering
		\caption{Comparison (PSNR-C / PSNR-S) with different gradients of rounding function.}
			\begin{tabular}{cccc}
			\toprule
			gradient & Round & JPEG Q=90 & JPEG Q=80         \\ \midrule
			all 0 & 78.32 / 11.76 & 82.19 / 11.76 & 81.98 / 11.76   \\
			all 1 & 41.15 / 40.28 & 31.46 / 29.85 & 29.48 / 27.91     \\
			GAF & 41.39 / 40.71 & 31.36 / 29.76 & 29.51 / 28.01   \\
			\bottomrule
			\label{round}
		\end{tabular}
\end{table}

{\bf The influence of $\lambda_c$ and $\lambda_s$.} Table \ref{lambda} illustrates how PSNR-C and PSNR-S vary inversely with each other. A higher proportion of $\lambda_c$ leads to a higher PSNR-C but a lower PSNR-S, while a lower proportion of $\lambda_c$ results in a lower PSNR-C but a higher PSNR-S. This implies that in practical applications, we can adjust the ratio of these two parameters according to different tasks.
\begin{table}[!htbp]
\centering
\caption{Performance comparison with different $\lambda_c$ and $\lambda_s$ under Gaussian $\sigma=10$.}

		\begin{tabular}{cccccc}
		\toprule
		$(\lambda_c$-$\lambda_s)$ & (0.1-1.9) & (0.5-1.5) & (1.0-1.0) & (1.5-0.5) & (1.9-0.1)          \\ \midrule
		PSNR-C & 23.61 & 27.66 & 31.13 & 34.23 & 39.54  \\
		PSNR-S & 34.89 & 30.73 & 29.34 & 27.32 & 22.97  \\
		\bottomrule
\label{lambda}
\end{tabular}
\end{table}
\subsection{Comparison with other methods}
\begin{figure}[htbp]
	\centering
	\includegraphics[width=0.8\linewidth]{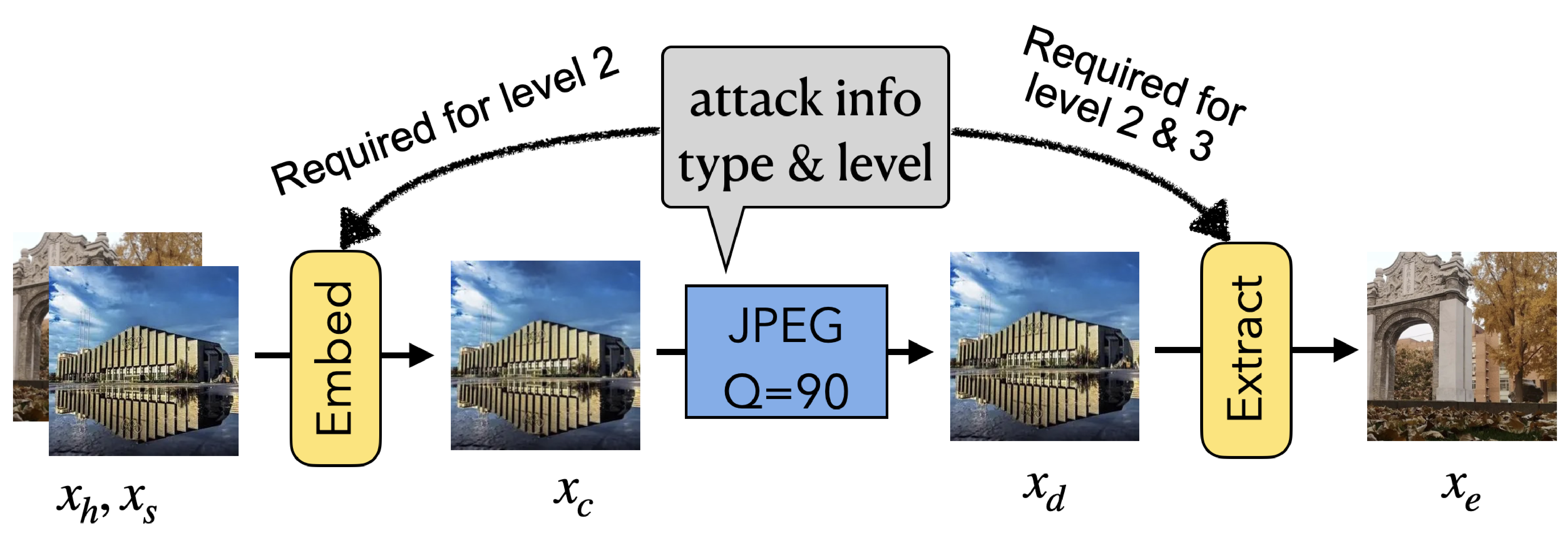}
	\caption{The difference between level 2 and 3. Level 2 requires the attack info both in embedding and extraction process, level 3 only need attack info in extraction process, which means level 2 needs to know which attack and the attack level before the attack actual happen.}
	\label{comparefig}
\end{figure}

 We categorize the task into four levels of difficulty and practicality, as shown below, and compare different methods across these levels. Fig. \ref{comparefig} illustrates the difference between level 2 and 3.

 {\bf Level 1:} Only deal with one specific attack method with one model.

 {\bf Level 2:} Deal with multiple attack methods with one model, but the attack information is required during both embedding and extraction.

 {\bf Level 3:} Deal with multiple attack methods with one model, but the attack information is only required during extraction.

 {\bf Level 4:} Deal with multiple attack methods with one model, and the attack information is not required.

We evaluate our PRIS method against several methods on different levels. To handle level 4 task, PRIS is trained for all attack methods with a 3-step strategy. To handle level 3 task, we take the trained PRIS model from level 4 task, and finetune different enhancement modules for different attack methods, i.e., the second step in the 3-step strategy. This means that in practice, we can select different enhancement modules for different attacks during the extraction process.

\begin{figure}[!htbp]
	\centering
	\includegraphics[width=0.7\linewidth]{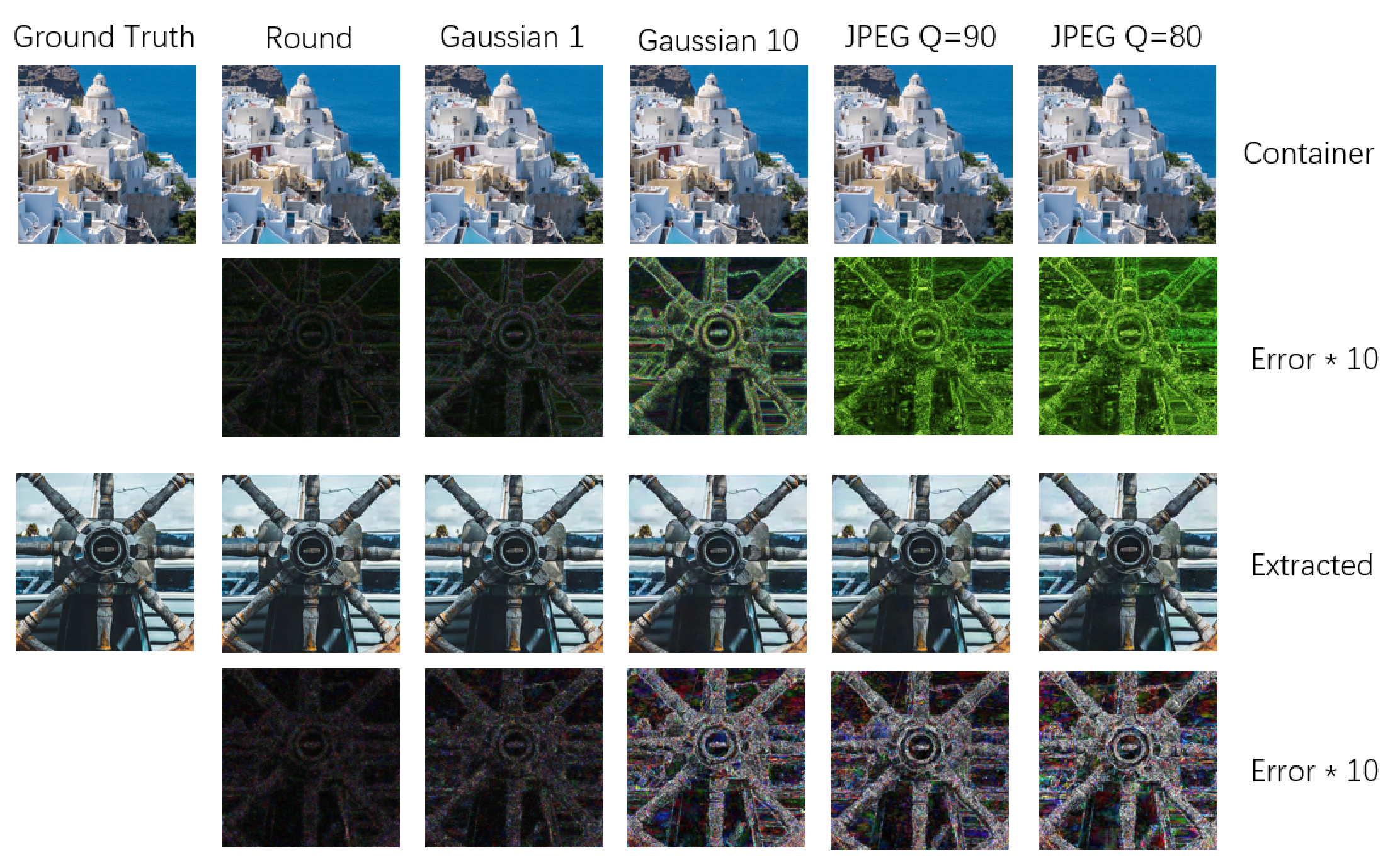}
	\caption{Visual results of PRIS in different type of attacks. }
	\label{visual}
\end{figure}

Table \ref{compare1} displays the performance results of PRIS and several other methods on the level 1 task. Our PRIS exhibits superior performance when compared to the latest methods, establishing it as the most robust model. Fig. \ref{visual} is the visual results of PRIS in different type of attacks on level 1, the cover-pair and secret-pair are undistinguished from human eyes under different distortions.

\begin{table}[!htbp]
	\caption{Comparison (PSNR-C / PSNR-S) between different methods on level 1. 
	}
	\centering
	\resizebox{\linewidth}{!}{
		\begin{tabular}{cccccc}
			\toprule
			Method & Gaussian $\sigma$=1 &Gaussian $\sigma$=10 & JPEG QF=90 &JPEG QF=80 & Round       \\ \midrule
			Baluja \cite{baluja2019hiding} & 21.94 / 20.64 & 21.36 / 19.86 & 21.86 / 20.14 & 21.53 / 19.81 & 22.05 / 20.68 \\
			HiNet \cite{jing2021hinet} & 37.69 / 36.70 & 30.79 / 28.99 & 31.28 / 29.60 & 29.34 / 27.66 & 40.45 / 39.78 \\
			ISN \cite{lu2021large}  & \ - \ / 28.98 & \ - \ / 27.12 & \ - \ / 27.48 & \ - \ / 27.15 & - \\
			RIIS \cite{RIIS}  & 29.67 / 30.39 & 27.77 / 28.11 & 28.17 / 28.53 & 27.73 / \textbf{28.19} & -           \\
			PRIS & \textbf{37.99 / 36.97} & \textbf{31.13 / 29.34} & \textbf{31.36 / 29.76} & \textbf{29.51} / 28.01 & \textbf{41.39 / 40.71} \\
			\bottomrule
	\end{tabular}}
	\label{compare1}
\end{table}

Table \ref{compare2} compares the results of RIIS (RIIS on level 2) and PRIS (level 4). Although level 3 requires less information than level 2, which means that level 3 is more challenging than level 2, PRIS still performance better than RIIS.

\begin{table}[!htbp]
	\caption{Comparison (PSNR-C / PSNR-S) between different methods, where RIIS and PRIS are trained and evaluated on level 2 and 3 respectively. 
}
	\centering
	\resizebox{\linewidth}{!}{
			\begin{tabular}{cccccc}
		\toprule
		Method & Gaussian $\sigma$=1   & Gaussian $\sigma$=10 & JPEG QF=90  & JPEG QF=80  & Round       \\ \midrule
		RIIS \cite{RIIS}  &  \ - \ / 30.01   & \ - \ / 28.03   & \ - \ / 28.44   & \ - \ / \textbf{28.10}   & -           \\
		PRIS & \textbf{29.69 / 33.32} &\textbf{29.69 / 28.65} &  \textbf{29.69 / 30.25}  &   \textbf{29.69} / 27.74  &   \textbf{29.69 / 33.70}  \\
		\bottomrule
	\end{tabular}}
	\label{compare2}
\end{table}

Table \ref{compare3} shows the results of HiNet, RIIS and PRIS on level 4 task. PRIS outperformed RIIS by 0.74 dB on PSNR-S in average. HiNet has the highest PSNR-C because it did not consider attacks during training, and the attacks occurred after the embedding process, so they did not affect PSNR-C. However, the PSNR-S of HiNet was unacceptable, especially for Gaussian $\sigma=10$ and JPEG distortions. Which indicates HiNet is fragile to attacks.

\begin{table}[!htbp]
	\caption{Comparison (PSNR-C / PSNR-S) between different methods on level 4.
}
	\centering
		\resizebox{\linewidth}{!}{
			\begin{tabular}{cccccc}
			\toprule
			Method & Gaussian $\sigma$=1 \   &\  Gaussian $\sigma$=10 \ &\  JPEG QF=90 \  & \ JPEG QF=80 \  &\  Round       \\ \midrule
			Baluja \cite{baluja2019hiding} & 22.12 / 20.95 & 22.12 / 20.1 & 22.12 / 20.45 & 22.12 / 19.96 & 22.12 / 20.96 \\
			HiNet \cite{jing2021hinet}  & \textbf{43.64} / 24.55 & \textbf{43.64} / 8.470  & \textbf{43.64} / 11.44 &\textbf{43.64} / 11.39 & \textbf{43.64 / 34.36}	\\
			ISN \cite{jing2021hinet}  &\ - \ / 25.19 & - / 8.55  &\  -\  / 11.25 & -  & -	\\
			RIIS \cite{RIIS}  &   \ - \ / 29.78   & \ - \ / 27.65    & \ - \ / 28.00   & \ - \ / \textbf{27.65}   & -           \\
			PRIS & 29.69 / \textbf{30.61} &    29.69 /  \textbf{28.43}  &     29.69 / \textbf{29.40} &   29.69 /  27.61 &   29.69 / 30.18  \\
			\bottomrule
		\end{tabular}}
	\label{compare3}
\end{table}

\subsection{Experiments in Practice}
Rounding is an inevitable distortion in practice, and it is the first attack will occur once the embedding is done. Thus, we applied rounding error prior to Gaussian or JPEG distortion respectively, which is more realistic, and labelled these two errors RGaussian and RJPEG. Furthermore, since the type of distortion that occurs when embedding the secret image is unpredictable, we trained our PRIS on level 3 and 4 with these realistic distortion. Table \ref{practical} shows that the PNSR-C is 30.3 dB and the lowest PNSR-S on level 3 and 4 are 27.00 and 26.79 dB respectively, which are satisfactory. We also deployed PRIS in \url{http://yanghang.site/hide/}, where users can embed a secret image into a host image and retrieve it by simply clicking the mouse.
\begin{table*}[!htbp]
	\centering
	\caption{Results (PSNR-C / PSNR-S) in practice.}
	\resizebox{\linewidth}{!}{
		\begin{tabular}{cccccc}
			\toprule
			Level & RGaussian $\sigma$=1& RGaussian $\sigma$=10 & RJPEG QF=90  & RJPEG QF=80   & Round   \\ \midrule
			3 & 30.30 / 35.94 & 30.30 / 28.33 &  30.30 / 29.81 &  30.30 / 27.00  &  30.30 / 36.94  \\
			4 & 30.30 / 34.48 & 30.30 / 28.03 & 30.30 / 28.91  &  30.30 / 26.79 &  30.30 / 35.11  \\
			\bottomrule
	\end{tabular}}
	\label{practical}
\end{table*}

\subsection{Test on Different Datasets}
In order to illustrate our robustness across different datasets, we evaluated our model, which was trained on DIV2K, directly on some additional datasets without finetune. Table \ref{differentdataset} shows our method is robust across different datasets. The description of additional datasets are as follows:
 
1.  ImageNet: This dataset contains 1,000 images, one randomly chosen from each class of the validation split of the ImageNet 2012 dataset.

2. COCO: This dataset contains 5,000 images from COCO 2017 validation split.

3. VOC : This dataset contains 2,510 images from VOC-2007 validation split. \\

\begin{table*}[!htbp]
	\caption{Comparison (PSNR-C / PSNR-S) between different datasets on level 1 with PRIS trained on DIV2K train split.}
	\centering
	\resizebox{\linewidth}{!}{
			\begin{tabular}{cccccc}
		\toprule
		Dataset &  Gaussian $\sigma$=1   &  Gaussian $\sigma$=10 &  JPEG QF=90  &  JPEG QF=80   & Round       \\ \midrule
		DIV2K  & 37.99 / 36.97 & 31.13 / 29.34 & 31.36 / 29.76 & 29.50 / 28.01 & 41.39 / 40.71           \\
		ImageNet & 36.98 / 35.79 & 30.41 / 28.28 & 29.97 / 28.20 & 28.50 / 26.52 & 39.90 / 39.26 \\
		COCO & 36.18 / 34.63 & 30.29 / 28.10 & 29.93 / 28.08 & 28.39 / 26.42 & 37.98 / 37.16 \\
		VOC & 36.94 / 35.77 & 30.22 / 28.13 & 29.66 / 27.97 & 28.23 / 26.30 & 39.64 / 39.35 \\
		\bottomrule
	\end{tabular}}
	\label{differentdataset}
\end{table*}
\section{Conclusion}
This paper presents a practical robust invertible network for image steganography, named PRIS. It incorporates two enhancement modules and a 3-step training strategy into invertible neural networks. Moreover, previous studies neglect the rounding distortion; however, it is unavoidable in practice, and its non-differentiability poses a challenge for training. Therefore, this paper proposes GAF to address this problem and applies rounding distortion before other distortions, which is more realistic. In addition, despite being trained on a resolution of 224 $\times$ 224 pixels, PRIS can be adapted to any resolution in practice. These features endow PRIS with the ability to outperform the existing SOTA method RIIS in terms of robustness and practicability.


\begin{thebibliography}{10}
	
	\bibitem{Kessler2011Overview}
	Gary~C. Kessler and Chet Hosmer.
	\newblock {\em An {Overview} of {Steganography}}, pages 51--107.
	\newblock Elsevier, 2011.
	
	\bibitem{Johnson1998Exploring}
	Neil~F. Johnson and Sushil Jajodia.
	\newblock Exploring steganography: Seeing the unseen.
	\newblock {\em Computer}, 31(2):26--34, 2 1998.
	
	\bibitem{lu2021large}
	Shao-Ping Lu, Rong Wang, Tao Zhong, and Paul~L Rosin.
	\newblock Large-capacity image steganography based on invertible neural
	networks.
	\newblock In {\em Proceedings of the IEEE/CVF Conference on Computer Vision and
		Pattern Recognition}, pages 10816--10825, 2021.
	
	\bibitem{kessler2004overview}
	Gary~C Kessler.
	\newblock An overview of steganography for the computer forensics examiner.
	\newblock {\em Forensic science communications}, 6(3):1--27, 2004.
	
	\bibitem{fridrich2011breaking}
	Jessica Fridrich, Jan Kodovsk{\`y}, Vojt{\v{e}}ch Holub, and Miroslav Goljan.
	\newblock Breaking hugo--the process discovery.
	\newblock In {\em Information Hiding: 13th International Conference, IH 2011,
		Prague, Czech Republic, May 18-20, 2011, Revised Selected Papers 13}, pages
	85--101. Springer, 2011.
	
	\bibitem{RIIS}
	Youmin Xu, Chong Mou, Yujie Hu, Jingfen Xie, and Jian Zhang.
	\newblock Robust invertible image steganography.
	\newblock In {\em Proceedings of the IEEE/CVF Conference on Computer Vision and
		Pattern Recognition}, pages 7875--7884, 2022.
	
	\bibitem{cheddad2010digital}
	Abbas Cheddad, Joan Condell, Kevin Curran, and Paul Mc~Kevitt.
	\newblock Digital image steganography: Survey and analysis of current methods.
	\newblock {\em Signal processing}, 90(3):727--752, 2010.
	
	\bibitem{barni2001improved}
	Mauro Barni, Franco Bartolini, and Alessandro Piva.
	\newblock Improved wavelet-based watermarking through pixel-wise masking.
	\newblock {\em IEEE transactions on image processing}, 10(5):783--791, 2001.
	
	\bibitem{fridrich2001detecting}
	Jessica Fridrich, Miroslav Goljan, and Rui Du.
	\newblock Detecting lsb steganography in color, and gray-scale images.
	\newblock {\em IEEE multimedia}, 8(4):22--28, 2001.
	
	\bibitem{hsu1999hidden}
	Chiou-Ting Hsu and Ja-Ling Wu.
	\newblock Hidden digital watermarks in images.
	\newblock {\em IEEE Transactions on image processing}, 8(1):58--68, 1999.
	
	\bibitem{lerch2016unsupervised}
	Daniel Lerch-Hostalot and David Meg{\'\i}as.
	\newblock Unsupervised steganalysis based on artificial training sets.
	\newblock {\em Engineering Applications of Artificial Intelligence}, 50:45--59,
	2016.
	
	\bibitem{luo2010edge}
	Weiqi Luo, Fangjun Huang, and Jiwu Huang.
	\newblock Edge adaptive image steganography based on lsb matching revisited.
	\newblock {\em IEEE Transactions on information forensics and security},
	5(2):201--214, 2010.
	
	\bibitem{ruanaidh1996phase}
	JJKO Ruanaidh, William~J Dowling, and Francis~M Boland.
	\newblock Phase watermarking of digital images.
	\newblock In {\em Proceedings of 3rd IEEE International Conference on Image
		Processing}, volume~3, pages 239--242. IEEE, 1996.
	
	\bibitem{kadhim2019comprehensive}
	Inas~Jawad Kadhim, Prashan Premaratne, Peter~James Vial, and Brendan Halloran.
	\newblock Comprehensive survey of image steganography: Techniques, evaluations,
	and trends in future research.
	\newblock {\em Neurocomputing}, 335:299--326, 2019.
	
	\bibitem{Baluja2017Hiding}
	Shumeet Baluja.
	\newblock Hiding {Images} in {Plain} {Sight}: Deep {Steganography}.
	\newblock 2017.
	
	\bibitem{zhang2020udh}
	Chaoning Zhang, Philipp Benz, Adil Karjauv, Geng Sun, and In~So Kweon.
	\newblock Udh: Universal deep hiding for steganography, watermarking, and light
	field messaging.
	\newblock {\em Advances in Neural Information Processing Systems},
	33:10223--10234, 2020.
	
	\bibitem{baluja2019hiding}
	Shumeet Baluja.
	\newblock Hiding images within images.
	\newblock {\em IEEE transactions on pattern analysis and machine intelligence},
	42(7):1685--1697, 2019.
	
	\bibitem{duan2019reversible}
	Xintao Duan, Kai Jia, Baoxia Li, Daidou Guo, En~Zhang, and Chuan Qin.
	\newblock Reversible image steganography scheme based on a u-net structure.
	\newblock {\em IEEE Access}, 7:9314--9323, 2019.
	
	\bibitem{duan2020high}
	Xintao Duan, Liu Nao, Gou Mengxiao, Dongli Yue, Zimei Xie, Yuanyuan Ma, and
	Chuan Qin.
	\newblock High-capacity image steganography based on improved fc-densenet.
	\newblock {\em IEEE Access}, 8:170174--170182, 2020.
	
	\bibitem{duan2020highx}
	Xintao Duan, Mengxiao Gou, Nao Liu, Wenxin Wang, and Chuan Qin.
	\newblock High-capacity image steganography based on improved xception.
	\newblock {\em Sensors}, 20(24):7253, 2020.
	
	\bibitem{jing2021hinet}
	Junpeng Jing, Xin Deng, Mai Xu, Jianyi Wang, and Zhenyu Guan.
	\newblock Hinet: Deep image hiding by invertible network.
	\newblock In {\em Proceedings of the IEEE/CVF International Conference on
		Computer Vision}, pages 4733--4742, 2021.
	
	\bibitem{byrnes2021data}
	Olivia Byrnes, Wendy La, Hu~Wang, Congbo Ma, Minhui Xue, and Qi~Wu.
	\newblock Data hiding with deep learning: A survey unifying digital
	watermarking and steganography.
	\newblock {\em arXiv preprint arXiv:2107.09287}, 2021.
	
	\bibitem{wan2022comprehensive}
	Wenbo Wan, Jun Wang, Yunming Zhang, Jing Li, Hui Yu, and Jiande Sun.
	\newblock A comprehensive survey on robust image watermarking.
	\newblock {\em Neurocomputing}, 2022.
	
	\bibitem{chan2004hiding}
	Chi-Kwong Chan and Lee-Ming Cheng.
	\newblock Hiding data in images by simple lsb substitution.
	\newblock {\em Pattern recognition}, 37(3):469--474, 2004.
	
	\bibitem{tamimi2013hiding}
	Abdelfatah~A Tamimi, Ayman~M Abdalla, and Omaima Al-Allaf.
	\newblock Hiding an image inside another image using variable-rate
	steganography.
	\newblock {\em International Journal of Advanced Computer Science and
		Applications (IJACSA)}, 4(10), 2013.
	
	\bibitem{hawi2004steganalysis}
	Tariq~Al Hawi, MA~Qutayri, and Hassan Barada.
	\newblock Steganalysis attacks on stego-images using stego-signatures and
	statistical image properties.
	\newblock In {\em 2004 IEEE Region 10 Conference TENCON 2004.}, pages 104--107.
	IEEE, 2004.
	
	\bibitem{zhi2003lsb}
	Li~Zhi, Sui~Ai Fen, and Yang~Yi Xian.
	\newblock A lsb steganography detection algorithm.
	\newblock In {\em 14th IEEE Proceedings on Personal, Indoor and Mobile Radio
		Communications, 2003. PIMRC 2003.}, volume~3, pages 2780--2783. IEEE, 2003.
	
	\bibitem{pan2011image}
	Feng Pan, Jun Li, and Xiaoyuan Yang.
	\newblock Image steganography method based on pvd and modulus function.
	\newblock In {\em 2011 International Conference on Electronics, Communications
		and Control (ICECC)}, pages 282--284. IEEE, 2011.
	
	\bibitem{tsai2009reversible}
	Piyu Tsai, Yu-Chen Hu, and Hsiu-Lien Yeh.
	\newblock Reversible image hiding scheme using predictive coding and histogram
	shifting.
	\newblock {\em Signal processing}, 89(6):1129--1143, 2009.
	
	\bibitem{nguyen2006multi}
	Bui~Cong Nguyen, Sang~Moon Yoon, and Heung-Kyu Lee.
	\newblock Multi bit plane image steganography.
	\newblock In {\em Digital Watermarking: 5th International Workshop, IWDW 2006,
		Jeju Island, Korea, November 8-10, 2006. Proceedings 5}, pages 61--70.
	Springer, 2006.
	
	\bibitem{imaizumi2014multibit}
	Shoko Imaizumi and Kei Ozawa.
	\newblock Multibit embedding algorithm for steganography of palette-based
	images.
	\newblock In {\em Image and Video Technology: 6th Pacific-Rim Symposium, PSIVT
		2013, Guanajuato, Mexico, October 28-November 1, 2013. Proceedings 6}, pages
	99--110. Springer, 2014.
	
	\bibitem{zhu2018hidden}
	Jiren Zhu, Russell Kaplan, Justin Johnson, and Li~Fei-Fei.
	\newblock Hidden: Hiding data with deep networks.
	\newblock In {\em Proceedings of the European conference on computer vision
		(ECCV)}, pages 657--672, 2018.
	
	\bibitem{zhang2019steganogan}
	Kevin~Alex Zhang, Alfredo Cuesta-Infante, Lei Xu, and Kalyan Veeramachaneni.
	\newblock Steganogan: High capacity image steganography with gans.
	\newblock {\em arXiv preprint arXiv:1901.03892}, 2019.
	
	\bibitem{shi2018ssgan}
	Haichao Shi, Jing Dong, Wei Wang, Yinlong Qian, and Xiaoyu Zhang.
	\newblock Ssgan: Secure steganography based on generative adversarial networks.
	\newblock In {\em Advances in Multimedia Information Processing--PCM 2017: 18th
		Pacific-Rim Conference on Multimedia, Harbin, China, September 28-29, 2017,
		Revised Selected Papers, Part I 18}, pages 534--544. Springer, 2018.
	
	\bibitem{dinh2014nice}
	Laurent Dinh, David Krueger, and Yoshua Bengio.
	\newblock Nice: Non-linear independent components estimation.
	\newblock {\em arXiv preprint arXiv:1410.8516}, 2014.
	
	\bibitem{zhu2017unpaired}
	Jun-Yan Zhu, Taesung Park, Phillip Isola, and Alexei~A Efros.
	\newblock Unpaired image-to-image translation using cycle-consistent
	adversarial networks.
	\newblock In {\em Proceedings of the IEEE international conference on computer
		vision}, pages 2223--2232, 2017.
	
	\bibitem{ardizzone2018analyzing}
	Lynton Ardizzone, Jakob Kruse, Sebastian Wirkert, Daniel Rahner, Eric~W
	Pellegrini, Ralf~S Klessen, Lena Maier-Hein, Carsten Rother, and Ullrich
	K{\"o}the.
	\newblock Analyzing inverse problems with invertible neural networks.
	\newblock {\em arXiv preprint arXiv:1808.04730}, 2018.
	
	\bibitem{song2019mintnet}
	Yang Song, Chenlin Meng, and Stefano Ermon.
	\newblock Mintnet: Building invertible neural networks with masked
	convolutions.
	\newblock {\em Advances in Neural Information Processing Systems}, 32, 2019.
	
	\bibitem{chen2019residual}
	Ricky~TQ Chen, Jens Behrmann, David~K Duvenaud, and J{\"o}rn-Henrik Jacobsen.
	\newblock Residual flows for invertible generative modeling.
	\newblock {\em Advances in Neural Information Processing Systems}, 32, 2019.
	
	\bibitem{kingma2018glow}
	Durk~P Kingma and Prafulla Dhariwal.
	\newblock Glow: Generative flow with invertible 1x1 convolutions.
	\newblock {\em Advances in neural information processing systems}, 31, 2018.
	
	\bibitem{behrmann2019invertible}
	Jens Behrmann, Will Grathwohl, Ricky~TQ Chen, David Duvenaud, and
	J{\"o}rn-Henrik Jacobsen.
	\newblock Invertible residual networks.
	\newblock In {\em International Conference on Machine Learning}, pages
	573--582. PMLR, 2019.
	
	\bibitem{jia2023afcihnet}
	Xingwang Jia, Huamei Xin, Lingchen Gu, Hao Wang, Jiande Sun, and Wenbo Wan.
	\newblock Afcihnet: Attention feature-constrained network for single image
	information hiding.
	\newblock {\em Engineering Applications of Artificial Intelligence},
	126:107105, 2023.
	
	\bibitem{huang2017densely}
	Gao Huang, Zhuang Liu, Laurens Van Der~Maaten, and Kilian~Q Weinberger.
	\newblock Densely connected convolutional networks.
	\newblock In {\em Proceedings of the IEEE conference on computer vision and
		pattern recognition}, pages 4700--4708, 2017.
	
	\bibitem{setiadi2021psnr}
	De~Rosal Igantius~Moses Setiadi.
	\newblock Psnr vs ssim: imperceptibility quality assessment for image
	steganography.
	\newblock {\em Multimedia Tools and Applications}, 80(6):8423--8444, 2021.
	
	\bibitem{agustsson2017ntire}
	Eirikur Agustsson and Radu Timofte.
	\newblock Ntire 2017 challenge on single image super-resolution: Dataset and
	study.
	\newblock In {\em Proceedings of the IEEE conference on computer vision and
		pattern recognition workshops}, pages 126--135, 2017.
	
\end{thebibliography}
\end{document}